\pdfoutput=1
\documentclass[sigconf]{acmart}
\usepackage{booktabs}
\usepackage{subfigure}
\usepackage{enumitem}
\usepackage{amsmath,amsfonts}
\usepackage{amssymb}
\usepackage{bm}
\usepackage{multirow}
\usepackage{booktabs}
\usepackage{algorithm}
\usepackage{algorithmic}
\def\method{KGNN}

\AtBeginDocument{%
  \providecommand\BibTeX{{%
    \normalfont B\kern-0.5em{\scshape i\kern-0.25em b}\kern-0.8em\TeX}}}

\copyrightyear{2022}
\acmYear{2022}
\setcopyright{acmcopyright}
\acmConference[WSDM '22] {Proceedings of the Fifteenth ACM International Conference on Web Search and Data Mining}{February 21--25, 2022}{Tempe, AZ, USA.}
\acmBooktitle{Proceedings of the Fifteenth ACM International Conference on Web Search and Data Mining (WSDM '22), February 21--25, 2022, Tempe, AZ, USA}
\acmPrice{15.00}
\acmISBN{978-1-4503-9132-0/22/02}
\acmDOI{10.1145/3488560.3498429}

\begin{document}
\fancyhead{}

\title{KGNN: Harnessing Kernel-based Networks \\ for Semi-supervised Graph Classification}


\author{Wei Ju}
\affiliation{%
  \institution{School of Computer Science,\\
    Peking University}
  \streetaddress{}
  \city{}
  \state{}
  \country{}}
\email{juwei@pku.edu.cn}

\author{Junwei Yang}
\affiliation{%
  \institution{School of Computer Science,\\
    Peking University}
  \streetaddress{}
  \city{}
  \state{}
  \country{}}
\email{yjwtheonly@pku.edu.cn}

\author{Meng Qu}
\affiliation{%
  \institution{Mila - Québec AI Institute,\\
  Université de Montréal}
  \streetaddress{}
  \city{}
  \state{}
  \country{}}
\email{meng.qu@umontreal.ca}

\author{Weiping Song}
\affiliation{%
  \institution{School of Computer Science,\\
    Peking University}
  \streetaddress{}
  \city{}
  \state{}
  \country{}}
\email{weiping.song@pku.edu.cn}

\author{Jianhao Shen}
\affiliation{%
  \institution{School of Computer Science,\\
    Peking University}
  \streetaddress{}
  \city{}
  \state{}
  \country{}}
\email{jhshen@pku.edu.cn}

\author{Ming Zhang}
\authornote{Corresponding author}
\affiliation{%
  \institution{School of Computer Science,\\
    Peking University}
  \streetaddress{}
  \city{}
  \state{}
  \country{}}
\email{mzhang\_cs@pku.edu.cn}

\begin{abstract}
This paper studies semi-supervised graph classification, which is an important problem with various applications in social network analysis and bioinformatics. This problem is typically solved by using graph neural networks (GNNs), which yet rely on a large number of labeled graphs for training and are unable to leverage unlabeled graphs. We address the limitations by proposing the Kernel-based Graph Neural Network (\method{}). A KGNN consists of a GNN-based network as well as a kernel-based network parameterized by a memory network. The GNN-based network performs classification through learning graph representations to \emph{implicitly} capture the similarity between query graphs and labeled graphs, while the kernel-based network uses graph kernels to \emph{explicitly} compare each query graph with all the labeled graphs stored in a memory for prediction. The two networks are motivated from complementary perspectives, and thus combing them allows \method{} to use labeled graphs more effectively. We jointly train the two networks by maximizing their agreement on unlabeled graphs via posterior regularization, so that the unlabeled graphs serve as a bridge to let both networks mutually enhance each other. 
Experiments on a range of well-known benchmark datasets demonstrate that \method{} achieves impressive performance over competitive baselines.
\end{abstract}

% <concept_id>10002950.10003624.10003633.10010917</concept_id>
% <concept_desc>Mathematics of computing~Graph algorithms</concept_desc>
% <concept_significance>300</concept_significance>
% <concept_id>10010147.10010257.10010293.10010294</concept_id>
% <concept_desc>Computing methodologies~Neural networks</concept_desc>
% <concept_significance>300</concept_significance>

% \ccsdesc[300]{Mathematics of computing~Graph algorithms}
% \ccsdesc[300]{Computing methodologies~Neural networks}


\maketitle

\section{Introduction}

Graph-structured data are ubiquitous in a wide range of domains. Examples include social networks \cite{song2019session}, biological reaction networks \cite{pavlopoulos2011using}, molecules \cite{sun2020infograph}. For graph-structured data, one fundamental problem is graph classification, which aims at analyzing and predicting the property of the entire graph. Such a problem has various downstream applications, including predicting the properties of molecules \cite{hao2020asgn} and analyzing the functionality of compounds \cite{kojima2020kgcn}.

Graph classification is typically formalized as a supervised learning task, and many recent works propose to use graph neural networks (GNNs) \cite{ying2018hierarchical,lee2019self,li2019semi} to solve the problem. The basic idea is to learn effective graph representations with nonlinear message passing schemas. At each step, a node receives messages from all its neighbors, which are further aggregated to update the node representation. Finally, a readout function is applied to integrate all the node representations into a representation of the whole graph. With this shared message passing framework, the learned graph representations can implicitly capture the similarity between query graphs and labeled graphs in the latent space. Despite the good performance, GNNs usually require a large amount of labeled data 
for training and fail to leverage unlabeled data. However, data annotation often requires domain experts, which is highly expensive, especially in specific domains such as biomedicine~\cite{hao2020asgn}.

This motivates us to study semi-supervised graph classification, i.e., using both labeled and unlabeled data for graph classification. The unlabeled data serve as a regularizer, which helps a model better explore the inherent graph semantic information even with a limited amount of labeled data. Indeed, there are a handful of works along this line~\cite{li2019semi,sun2020infograph,hao2020asgn}, and they typically employ semi-supervised learning techniques to train GNN models. These approaches usually integrate self-training~\cite{lee2013pseudo} or knowledge distillation~\cite{hinton2015distilling} into GNNs. 
However, these methods suffer from two key limitations:
\textbf{(1) Unable to well explore graph similarity}. Graph classification relies on comparing the query graphs with labeled graphs. Existing methods~\cite{li2019semi,sun2020infograph,hao2020asgn} typically compute the similarity of graphs in an implicit way by projecting graphs into a latent space with a GNN encoder. However, such implicit methods often cannot well explore the similarity of graphs. \textbf{(2) Suffering from labeled data scarcity}. Besides, experimental results show that the performance of existing methods is still unsatisfactory especially when labeled data are very scarce. The reason is that these methods are not able to obtain high-quality annotated data to improve model training. Therefore, we are looking for an approach that is able to better consider the relationship among graphs and meanwhile overcome the challenge of scarce labeled data.

In this paper, we propose such a method called the Kernel-based Graph Neural Network (\method{}). The key idea of \method{} is to enhance GNNs with graph kernels, which are able to explicitly measure the graph similarity of graphs. 
To leverage graph kernels effectively, we introduce two modules in a KGNN, i.e., a GNN-based network and a kernel-based network. The GNN-based network is parameterized by existing GNNs, which uses message passing mechanisms to learn useful graph representations for graph classification. In contrast, the kernel-based network employs a memory network, where the memory stores the given labeled graphs. Given a query graph, we compare it with all the graphs in the memory using graph kernels, and further integrate the labels of the most similar graphs to predict the label of the query graph.

The GNN-based network and kernel-based network explore graph similarity from different angles, i.e., message passing and graph kernels respectively. Although they are naturally complementary, how to jointly train both networks which enables us to distill the knowledge between each other is nontrivial. We solve the problem with a novel posterior regularization framework \cite{ganchev2010posterior}, which encourages both networks to collaborate with each other and maximize their agreement on unlabeled data. Each training iteration of posterior regularization consists of two steps. In the first step, the kernel-based network is updated by projecting the GNN-based network into a regularized subspace, yielding a stronger kernel-based network. In the second, we distill the knowledge learned by the kernel-based network into the GNN-based network, so that allowing the GNN to better explore graph similarity and overcome the challenge caused by the scarcity of labeled data.

To summarize, the main contributions of this work are as follows:
\begin{itemize}
\item We propose a novel approach for semi-supervised graph classification, which consists of a GNN-based network and a kernel-based network to fully capture the graph similarity and overcome the scarcity of labeled data.
\item We develop a novel posterior regularization framework to combine the advantages of graph neural networks and graph kernels in a principled way, such that they can mutually enhance each other via optimizing the two modules with an EM-style algorithm.
\item We conduct extensive experiments on a range of well-known benchmarks to demonstrate the effectiveness of our \method{}.
\end{itemize}

\section{Related Work}
\label{sec::related}

\subsection{Graph Neural Networks}

Our work is related to graph neural networks (GNNs), which are able to learn useful graph representations via supervised learning. Most existing methods~\cite{xu2019powerful,ying2018hierarchical,lee2019self,khasahmadi2020memory} inherently use a message-passing neural network \cite{gilmer2017neural} to learn node representations, and then aggregate them as graph representations.
Because of their effectiveness in learning graph representations, they achieve state-of-the-art results. However, graph representations derived from GNNs often fail to effectively leverage graph similarity for prediction. Besides, these methods rely on a large number of labeled data for training.
With \method{}, besides learning effective graph representations derived from GNNs, we also benefit from graph kernels to explicitly incorporate graph similarity in a semi-supervised framework.

%0.5\textwidth
\begin{figure}[t]
	\centering
	\includegraphics[width=1.0\linewidth]{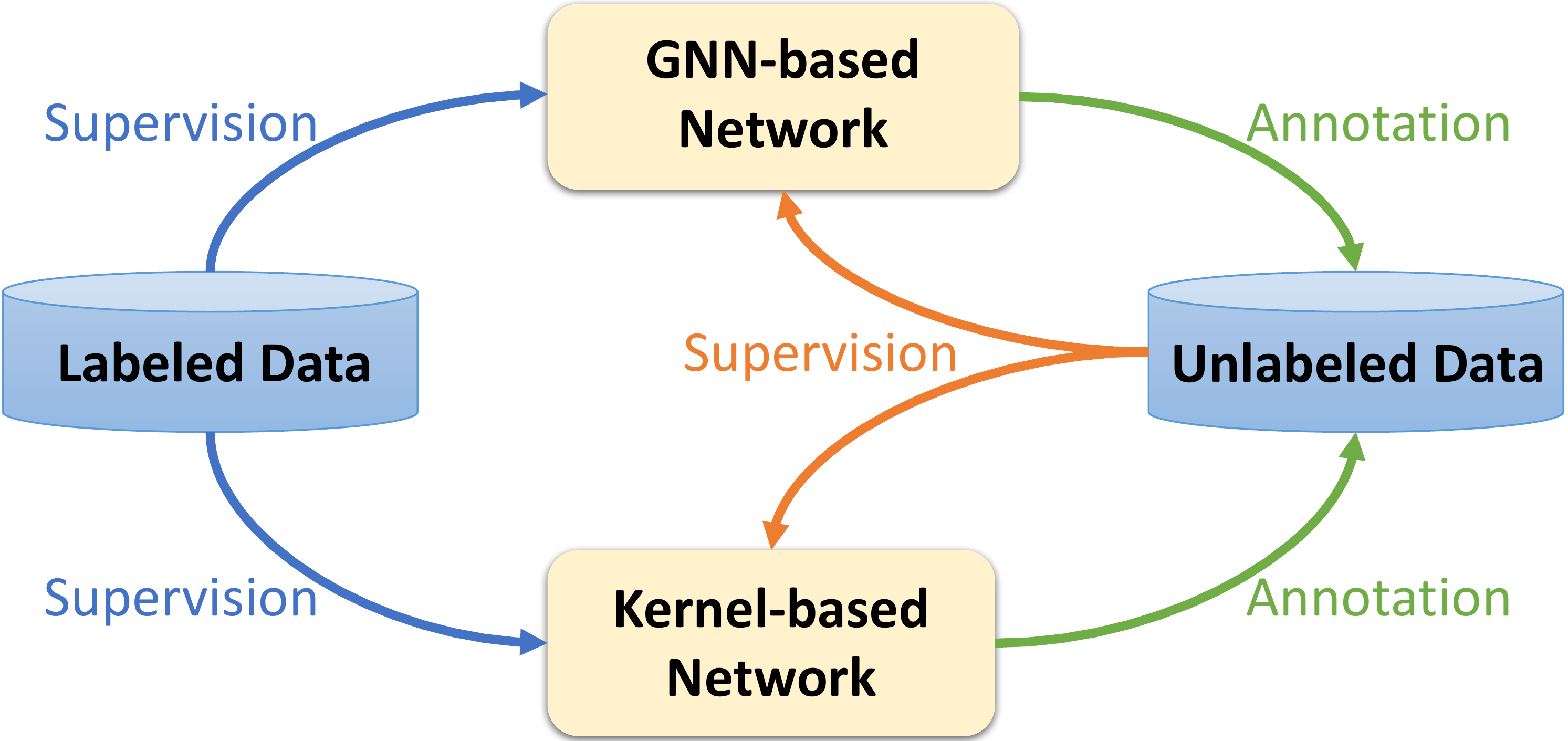}
	\caption{Illustration of the framework. \method{} consisting of a GNN-based network and a kernel-based network receives supervision from labeled data to annotate the unlabeled data, which improve both modules in turn. Two modules are jointly optimized and mutually enhance each other.}
	\label{fig::illustration}
\end{figure}

\subsection{Graph Kernels}
% and up-to-date
Kernels on graphs are originally introduced by \cite{gartner2003graph,kashima2003marginalized}, 
and the basic idea is to decompose graphs into atomic substructures and use kernel functions to capture the graph similarity, which can be used in kernel methods for graph classification. 
Many early methods\cite{gartner2003graph,kashima2003marginalized} define feature vectors as counts of label paths using a random walk or shortest path. However, these methods cannot be scaled to large graphs and suffer from high computational complexity. To alleviate these weaknesses, many later graph kernels are designed based on limited-sized substructures ~\cite{shervashidze2009efficient} or neighborhood aggregation~\cite{shervashidze2011weisfeiler}. Some recent studies also try to combine graph kernels and graph neural networks~\cite{du2019graph,chen2020convolutional}. GCKNs \cite{chen2020convolutional} connect two methods via multilayer graph kernels. GNTKs \cite{du2019graph} prove that their kernels are equivalent to infinitely wide GNNs trained by gradient descent. Similar to these methods, our proposed \method{} also combines the advantages of both worlds but from different views.
To be specific, our approach not only imposes graph kernels as structured constraints to guide the training process of GNNs, but also uses unlabeled data to assist model training by maximizing the agreement of the two networks on unlabeled data, while their works can only leverage labeled data in a supervised way.

\subsection{Semi-supervised Learning}
Another category of related work is semi-supervised learning. The self-training method is one of the earliest ideas along this line. It requires a trained classifier to periodically predict class labels of the unlabeled data and add high-quality classified samples to the training set. For example, the entropy minimization method \cite{grandvalet2005semi} requires the classifier to output low-entropy predictions on unlabeled data. Another line of research focuses on perturbation-based approaches. The core idea is to encourage the mapping consistency between the input and output when noise is applied to the input, which is so-called consistency regularization. Among them, the temporal ensembling model \cite{laine2017temporal} maintains an exponential moving average of label predictions while the mean teacher model \cite{tarvainen2017mean} averages network parameters to obtain a stable target output.

Besides, there are also some recent studies \cite{li2019semi,sun2020infograph,hao2020asgn} proposed for semi-supervised graph classification. 
SEAL-AI \cite{li2019semi} approaches the problem in the perspective of the hierarchical graph related to self-training. InfoGraph \cite{sun2020infograph} and ASGN \cite{hao2020asgn} 
learn graph representations via contrastive learning and active learning respectively, which both can be boiled down to the teacher-student framework.
Compared with existing works, our work focuses on combining the advantages of graph neural networks and graph kernels via posterior regularization, incorporating graph similarity from implicit and explicit perspectives respectively, while their works fail to enhance GNNs from the view of topological similarity.

\section{Problem Definition \& Preliminary}
\label{sec::definition}
\subsection{Problem Definition}

\smallskip
\noindent\textbf{Definition: Semi-supervised Graph Classification. } Let $G = (V, E, \mathbf{X})$ represent a graph, where $V$ is the set of nodes, $E$ is the set of edges. We use $\mathbf{X}\in R^{|V|\times d}$ denotes the node feature matrix, where $d$ is the dimension of features. For semi-supervised graph classification, let $\mathcal{G} = \{\mathcal{G}^L, \mathcal{G}^U\}$ denote a set of graphs, in which $\mathcal{G}^L = \left\{G_{1}, \cdots, G_{\left|\mathcal{G}^{L}\right|}\right\}$ are labeled graphs and $\mathcal{G}^{U}=$ $\left\{G_{\left|\mathcal{G}^{L}\right|+1}, \cdots, G_{\left|\mathcal{G}^{L}\right|+\left|\mathcal{G}^{U}\right|}\right\}$ are unlabeled graphs. Additionally, we use $\mathbf{y}^L$ to represent the labels corresponding to $\mathcal{G}^L$. The goal of semi-supervised graph classification is to learn a label distribution $p(\mathbf{y}^U |\mathcal{G}, \mathbf{y}^L) $, which can assign labels to unlabeled graphs $\mathcal{G}^U$.

\subsection{Graph Kernels}
\label{sec::gk}

Given two graphs $G_{i}$ and $G_{j}$, the graph kernel $K(G_{i}, G_{j})$ measures the similarity between them and is defined as:
\begin{equation}
    K\left(G_{i}, G_{j}\right)=\sum_{u \in V_{i}} \sum_{v \in V_{j}} k_{base }\left(f_{G_{i}}\left(u\right), f_{G_{j}}\left(v\right)\right),
\end{equation}
where the base kernel $k_{base}$, i.e., the inner product on Hilbert space, compares substructures centered at nodes~$v_{i}$ and $v_{j}$, and $f_G(v)$ is the feature vector counting the number of appearances of each substructure (e.g., subtrees, paths, graphlets) in the graph $G$.

Afterward, the matrix of pairwise similarities among graphs defined by graph kernels can be passed to kernel-based methods such as the Support Vector Machine to perform graph classification.

\smallskip
\noindent\textbf{Weisfeiler-Lehman Kernel.}
Inspired by the Weisfeiler-Lehman test of graph isomorphism \cite{weisfeiler1968reduction}, Weisfeiler-Lehman kernel \cite{shervashidze2011weisfeiler} follows the idea of the iterative relabeling process to augment the label of each node through the sorted set of labels of its neighbors in each iteration. The augmented labels are further compressed into a new label. Then the final feature for each graph is the concatenation of label counts for each iteration.

\subsection{Posterior Regularization}
Posterior regularization (PR) \cite{ganchev2010posterior} is a principled framework to impose constraints from the desired distribution $q(\mathbf{y})$ on probabilistic models $p_{\theta}\left(\mathbf{y}|\mathbf{x}\right)$ as follows:
\begin{equation}
\label{eqn::PR}
\mathcal{L}(q;\bm{\theta})=\mathcal{L}(\bm{\theta})-\min _{q \in Q} \text{KL}\left(q(\mathbf{y}) \| p_{\theta}\left(\mathbf{y}|\mathbf{x}\right)\right),
\end{equation}
where $\mathcal{L}(\bm{\theta})$ is the log-likelihood of model $p_{\theta}\left(\mathbf{y}|\mathbf{x}\right)$, and $Q$ is a constraint set, with respect to the expectations of constraint features $\psi(\mathbf{x}, \mathbf{y})$ that are bounded by $\mathbf{b}$:
\begin{equation}
\label{eqn::constraint}
Q=\left\{q(\mathbf{y}): \mathbb{E}_{q}[\psi(\mathbf{x}, \mathbf{y})] \leq \mathbf{b}\right\}.
\end{equation}
To optimize the posterior regularized likelihood $\mathcal{L}(q;\bm{\theta})$, PR presents an iterative procedure akin to an EM-style algorithm:
% minimization-maximization

\begin{align}
&\text{E}: q^{t+1}=\underset{q \in Q}{\arg \min } \text{KL}(q(\mathbf{y}) \| p_{\theta^{t}}(\mathbf{y}|\mathbf{x})), \label{eqn::E-step}\\
&\text{M}: \bm{\theta}^{t+1}=\underset{\theta}{\arg \max } \mathcal{L}(\bm{\theta})+\mathbb{E}_{q^{t+1}}[\log p_{\theta}\left(\mathbf{y}|\mathbf{x}\right)]. \label{eqn::M-step}
\end{align}

For better interpretation, it can be viewed as performing coordinate ascent on $\mathcal{L}(q;\bm{\theta})$. Starting from an initial parameter estimate $\theta^0$, the algorithm iterates two block-coordinate ascent steps until a convergence criterion is reached.
\section{\method{}: Kernel-based Graph Neural Network}
\label{sec::model}
% In this section, we first present a description of the framework of \method{}. Then we describe the components of \method{} in detail.

\begin{figure}
	\centering
	\includegraphics[width=1.0\linewidth]{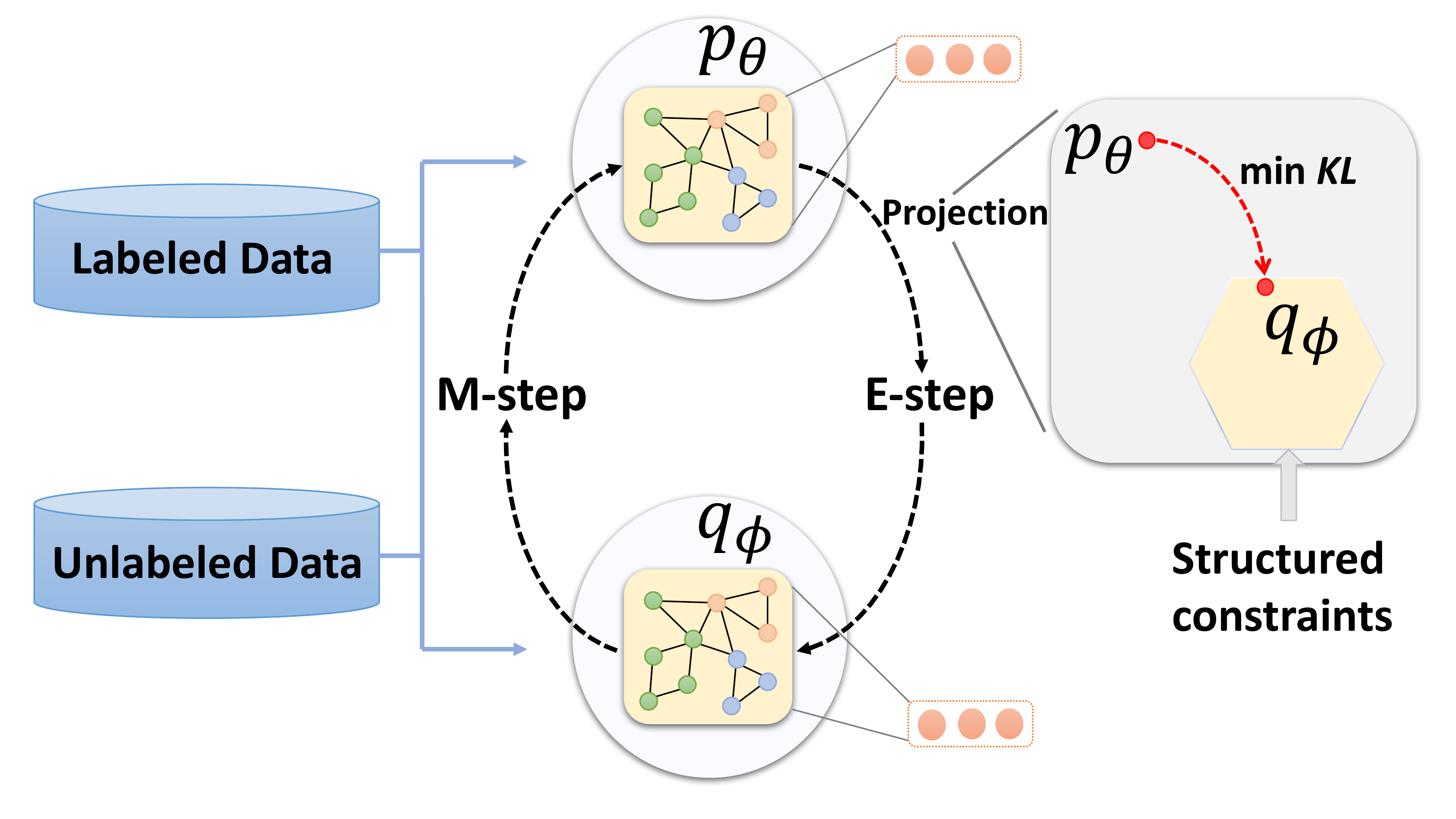}
% 	\vspace{-1mm}
	\caption{Framework overview. Yellow squares are the entire graph for classification. Orange triple circles are graph representations. \method{} consists of a GNN-based network $p_\theta$ and a kernel-based network $q_\phi$, which is trained by alternating between an E-step and an M-step. In the E-step, the kernel-based network $q_\phi$ is updated by projecting the GNN-based network $p_\theta$ to a regularized subspace (red dashed arrow). In the M-step, the GNN-based network $p_\theta$ is updated from the knowledge distilled by the kernel-based network $q_\phi$.}
	\label{fig::framework}
\end{figure}

\subsection{Overview}
\label{sec::overview}

In this paper, we introduce our approach for semi-supervised graph classification by incorporating both labeled and unlabeled data. Existing methods typically combine graph neural networks with semi-supervised learning techniques. However, these methods are not able to fully leverage graph similarity information for classification, which leads to insufficient expressiveness. Also, their performance is not yet satisfactory when the labeled data are limited.

We address the above limitations by proposing the Kernel-based Graph Neural Network (\method{}), which combines methods based on graph kernels and graph neural networks. The idea is to use graph kernels as structured constraints to guide the training of graph neural networks via posterior regularization, allowing the graph neural network to explicitly exploit graph similarity and receive additional supervision to boost performance. In particular, there are a kernel-based network and a GNN-based network in a KGNN, and we use an EM-style algorithm to jointly train both networks. In the E-step, the kernel-based network represented as a memory network is updated by projecting the GNN-based network to a subspace constrained by the graph similarity. In the M-step, the GNN-based network is improved by distilling the knowledge from the kernel-based network. In this way, both networks can effectively mutually enhance each other. An illustration of the framework is presented in Figure~\ref{fig::framework}. Next, we first introduce the GNN-based network and the kernel-based network.  Finally, the training algorithm is explained.

\subsection{GNN-based Network}
\label{sec::p}
Due to the appealing representation learning ability, graph neural networks have been widely used in graph classification tasks. For example, InfoGraph~\cite{sun2020infograph} first learns node representations with GIN~\cite{xu2019powerful}, and then summarizes the node representations into the graph-level representation, which serves as graph features for the classification task. Note that in InfoGraph~\cite{sun2020infograph}, the label dependency between different graphs is ignored. We follow the same idea and factorize the joint label distribution as:
\begin{equation}\label{eq::p_label_fact}
    p(\mathbf{y}^U |\mathcal{G}, \mathbf{y}^L) = \prod_{g\in \mathcal{G}^U} p(\mathbf{y}^g| \mathcal{G}, \mathbf{y}^L).
\end{equation}

Furthermore, to learn graph-level representation for a graph $G=(V, E, \mathbf{X})$, GNN-based methods solely rely on the node attributes $\mathbf{X}$ and graph structure $E$. In other words, it is independent of other graphs, we therefore define the conditional probability of label $\mathbf{y}^g$ for graph $G$ as follows:
\begin{equation}
    p(\mathbf{y}^g | \mathcal{G}, \mathbf{y}^L) = p(\mathbf{y}^g | \mathbf{X}, E).
\end{equation}

To predict the graph label $\mathbf{y}^g$, a graph neural network first learns node representations by aggregating and updating messages from its neighborhoods. Based on the node representations, a READOUT function summarizes them into the whole graph representation. This procedure can be formalized in the following way:

\begin{equation}
    \begin{gathered}
        \mathbf{h}_{v}^{(l)} = U_\theta^{(l)} \left( \mathbf{h}_{v}^{(l-1)}, A_\theta^{(l)} \left( \left\{ \mathbf{h}_{u}^{(l-1)} \right\}_{u \in \mathcal{N}(v)} \right)\right), \\
        H(G) = \text{READOUT} \left(\left\{\mathbf{h}_{v}^{L}\right\}_{v \in V}\right), \\
        p_\theta(\mathbf{y}^g | \mathbf{X}, E) = \text{MLP}_\theta \left( H(G) \right),
    \end{gathered}
\end{equation}
where $\mathbf{h}_{v}^{(l)}$ is the feature vector of node $v$ at the $l$-th layer. $\mathbf{h}_{v}^{(0)}$ is often initialized as node features (i.e., $\mathbf{X}_v$) and $\mathcal{N}(v)$ are neighborhoods to node $v$. The message-passing phase runs for $L$ steps and is defined in terms of aggregation functions $A_\theta^{(l)}(\cdot)$ and node updating functions $U_\theta^{(l)}(\cdot)$, which have a series of possible implementations \cite{kipf2017semi,hamilton2017inductive,velivckovic2018graph}. After a permutation invariant function $\text{READOUT}(\cdot)$ such as some pooling functions \cite{ying2018hierarchical,lee2019self,khasahmadi2020memory}, the final graph-level representation $H(G)$ is fed into an $\text{MLP}_\theta$ (Multi-Layer Perceptron) classifier for classification. In this paper, we follow InfoGraph~\cite{sun2020infograph} and use GIN \cite{xu2019powerful} to learn node representations for its high expressive power, and then use the sum operation to produce the graph-level representations (i.e., the READOUT function).

\begin{figure}[t]
	\centering
	\includegraphics[width=1.0\linewidth]{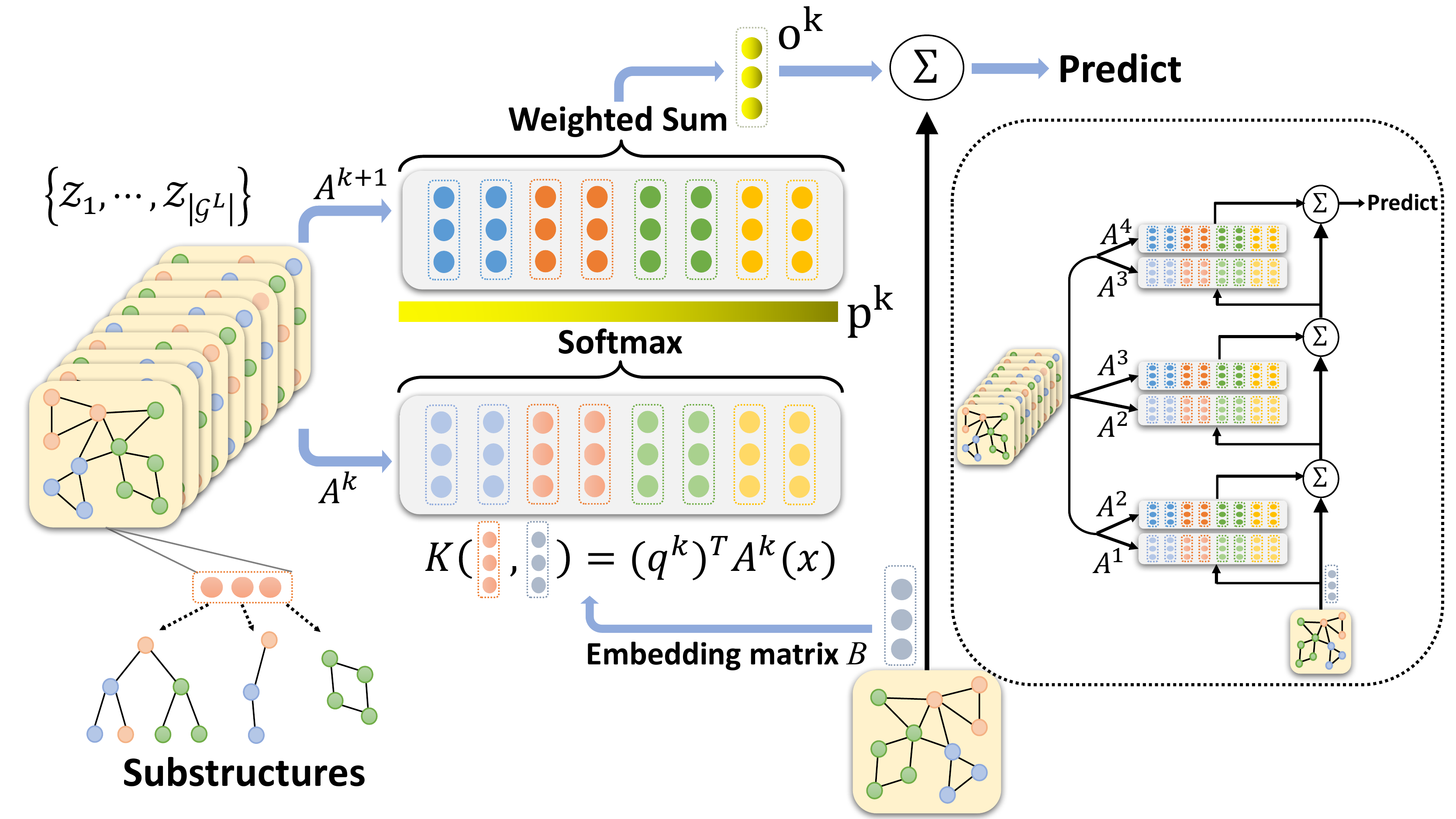}
	\caption{Illustration of the kernel-based network $q_\phi$. It is parameterized by a memory network that incorporates the graph kernels to capture the topological similarity information of graphs. The multi-hop mechanism helps in learning correlations between memories. For simplicity, we omit the subscript  $\phi$ of embedding matrix $A_\phi$ and $B_\phi$.}
	\label{fig::memory}
\end{figure}

\subsection{Kernel-based Network}
\label{sec::q}
Graph neural networks can effectively utilize node attributes and graph local structures under the message-passing framework, however, they are insufficient in capturing more global graph topology and ignore the relations between different graphs. This motivates us to additionally incorporate graph kernels, which have been proven useful for encoding graph topology and similarity information \cite{kriege2020survey}.

Traditional kernel-based methods perform graph classification tasks in a two-step approach that similarity features are separated from the training phase and are not optimized for downstream tasks, hence the performance is usually unsatisfactory. To address the above limitation, we propose a kernel-based network $q(\mathbf{y}^U|\mathcal{G}, \mathbf{y}^L)$ and implement it with a memory network, which shares a similar idea to classical kernel-based methods\cite{nowicki2010flexible}, but haves larger model capacity and can be optimized in an end-to-end manner. Moreover, memory networks can be naturally and effectively coupled with kernel tricks \cite{garcia2004hopfield} and introduced to further compare graph-graph similarity via graph kernels for prediction.

In detail, with the assumption of label dependency in Eq.~\ref{eq::p_label_fact}, we model the conditional probability $q(\mathbf{y}^g| \mathcal{G}, \mathbf{y}^L)$ as follows \begin{equation}
    q_\phi(\mathbf{y}^g | \mathcal{G}, \mathbf{y}^L) = \text{MemNN}_\phi (\mathcal{G}),
\end{equation}
where the memories of $\text{MemNN}_\phi$ consists of the labeled data set $\mathcal{G}^{L}$ represented as the feature vector set $\{z_{1}, \cdots, z_{\left|\mathcal{G}^{L}\right|}\}$ derived from graph kernels (see. Sec.~\ref{sec::gk}). Then each graph are converted into memory vectors by a set of trainable embedding matrix set $A_\phi = \{ A_\phi^1,\dots,A_\phi^{K+1} \}$, where each $A_\phi^k$ maps memory content into continuous space at each layer $k$. The query graph (one of the labeled data) is also encoded via another trainable embedding matrix $B_\phi$ to obtain a hidden representation $q^k$ as a reading head. Afterward, we measure the similarity between $q^k$ and each memory vector $A_\phi^k(z_i)$ by inner product, corresponding to the graph kernel computed on Hilbert space $\mathcal{H}$ and then followed by a softmax function. The model loops over $K$ hops and it computes the attention weights at hop $k$ for each memory $G_{i}$ using:
\begin{equation}
  p^k_i = \text{Softmax}((q^k)^{\top} A_\phi^k(z_i)),
\end{equation}
where $\text{Softmax}(x_i)=e^{x_i}/\Sigma_j {e^{x_j}}$. Here, $p^k$ is a soft memory selector that decides the memory relevance with respect to $q^k$. Then, the model reads out the memory $o^k$ by the weighted sum over $A_\phi^{k+1}$~\footnote{Here is $A_\phi^{k+1}$ since we use adjacent weighted tying  \cite{sukhbaatar2015end}.},
\begin{equation}
  o^k = \sum_i p^k_i A_\phi^{k+1}(z_i).
\end{equation}
Then, the query vector is updated for the next hop by using $q^{k+1} = q^{k} + o^{k}$. The above step leads to the final graph representation $o^K$, which will be fed into an $\text{MLP}_\phi$ classifier for prediction as follows:
\begin{equation}
    q_\phi(\mathbf{y}^g | \mathcal{G}, \mathbf{y}^L) = \text{MLP}_\phi \left( o^K \right).
\end{equation}
In this way, the representation of graphs can be combined with similarity features of multiple graphs from memories, and the multi-hop attention mechanism helps in learning correlations between memories. Thus, the derived kernel-based network enables high flexibility and is able to effectively exploit graph topology similarity. An illustration of the kernel-based network is presented in Figure~\ref{fig::memory}.

\subsection{Optimization Framework with PR}
Graph kernels offer effective prior knowledge to explicitly capture graph similarity. To inject this knowledge into graph neural networks efficiently, posterior regularization (PR) is utilized to provide a principled way to impose structured constraints on the posterior distribution of the probabilistic models. However, PR typically uses a fully-specified constraint (see Eq. (\ref{eqn::constraint})), which is often sub-optimal. To address this limitation, we represent the desired distribution $q(\mathbf{y})$  in (\ref{eqn::PR}) as the kernel-based network described in Section~\ref{sec::q}, where this manipulation leads to a more efficient and flexible training algorithm. For simplicity, the GNN-based network and the kernel-based network will be simply written $p_{\theta}(\mathbf{y}|\mathbf{x})$ and $q_{\phi}(\mathbf{y}|\mathbf{x})$ respectively with an abuse of notation in the rest of the paper.

Based on this, the kernel-based network enables learnable constraints and should be a good complement for the GNN-based network to incorporate the topology information. To this end, we formulate the training objective as:
\begin{equation}\label{eqn::goal}
\mathcal{L}(\bm{\theta, \bm{\phi}})=\mathcal{L}(\bm{\theta})-\min\nolimits_{\phi} \text{KL} \left(q_{\phi}(\mathbf{y}|\mathbf{x}) \| p_{\theta}\left(\mathbf{y}|\mathbf{x}\right)\right).
\end{equation}
Note that another benefit of modeling the kernel-based network as the memory network is the differentiability of our training objective Eq. (\ref{eqn::goal}). It is easy to use stochastic gradient descent algorithms to optimize model parameters. Next, we will give the specific training algorithm for updating parameters in Section~\ref{sec::training}.

\subsection{Training Algorithm}
\label{sec::training}
In the training phase, our goal is to maximize the posterior regularized likelihood (\ref{eqn::goal}) that can be optimized by an iterative procedure akin to an EM-style algorithm, alternated by an E-step and an M-step. In the E-step, the goal is to optimize the kernel-based network 
by minimizing the KL divergence between the kernel-based network and the GNN-based network. In the M-step, we optimize the GNN-based network by the original likelihood of model $p_{\theta}\left(\mathbf{y}|\mathbf{x}\right)$ and additional guidance derived from the kernel-based network. Next, we introduce the details of the E-step and M-step.

\begin{algorithm}[tb]
\caption{Joint Learning Algorithm of \method{}}
\label{alg::optim}
\begin{flushleft}
\textbf{Input:} Labeled data $\mathcal{G}^L$, unlabeled data $\mathcal{G}^U$. \\
\textbf{Parameter}: Model parameter $\bm{\theta}$ and $\bm{\phi}$ \\
\textbf{Output}: Jointly learned $p_{\theta}\left(\mathbf{y}|\mathbf{x}\right)$ and  $q_{\phi}\left(\mathbf{y}|\mathbf{x}\right)$.
\end{flushleft}

\begin{algorithmic}[1] %[1] enables line numbers
\STATE Initialize model parameter $\bm{\theta}$ on labeled data $\mathcal{G}^L$.
\WHILE{not convergence}
    \STATE \textrm{// E-step}\\
    \STATE Annotate unlabeled data $\mathcal{G}^U$ with $p_\theta$.
    \STATE Optimize model parameter $\bm{\phi}$ with Eq.~\eqref{eqn::obj-q}.

    \STATE \textrm{// M-step}\\
    \STATE Annotate unlabeled data $\mathcal{G}^U$ with $q_\phi$.
    \STATE Optimize model parameter $\bm{\theta}$ with Eq.~\eqref{eqn::obj-p}.
\ENDWHILE
\end{algorithmic}
\end{algorithm}

\smallskip
\noindent\textbf{E-step.}
\label{sec::E-step}
In this step, we optimize the parameter $\phi$. Recall that the optimization of the kernel-based network parameter $\phi$ is performed by minimizing the KL divergence in Eq. (\ref{eqn::E-step}). However, directly optimizing the likelihood function requires dealing with the partition function in $q_\phi$, which can be difficult. To address the restriction, we propose to instead minimize the reverse KL divergence:

\begin{equation}
\begin{aligned}
\label{eqn::obj-q-u}
    &\min\nolimits_\phi \text{KL}(p_\theta(\mathbf{y}|\mathbf{x}) \| q_\phi(\mathbf{y}|\mathbf{x}))\\
     = &\min\nolimits_\phi -\mathbb{E}_{p_\theta(\mathbf{y}|\mathbf{x})}[\log q_\phi(\mathbf{y}|\mathbf{x})] + const.
\end{aligned}
\end{equation}

Besides, we notice that $q_\phi$ is a standard discriminative classifier and can be also trained with labeled data. Therefore, we optimize $q_\phi$ by maximizing the following objective function:

\begin{equation}
\begin{aligned}
\label{eqn::obj-q}
\bm{\phi} = \underset{\phi}{\arg \max }  &\sum\nolimits_{(\mathbf{x},\mathbf{y}) \in L} \log q_\phi(\mathbf{y}|\mathbf{x}) \\
&+\mathbb{E}_{p_{\theta}(\mathbf{y}|\mathbf{x})}[\log q_\phi\left(\mathbf{y}|\mathbf{x}\right)].
\end{aligned}
\end{equation}

\smallskip
\noindent\textbf{M-step.}
\label{sec::M-step}
In this step, we optimize the parameter $\theta$. More specifically, besides optimizing the original likelihood of model $p_{\theta}$,
we will fix $q_\phi$ and further update $p_\theta$. Afterwards, the parameter $\theta$ can be updated by maximizing the following overall objective function:

\begin{equation}
\begin{aligned}
\label{eqn::obj-p}
\bm{\theta} = \underset{\theta}{\arg \max }  &\sum\nolimits_{(\mathbf{x},\mathbf{y}) \in L} \log p_\theta(\mathbf{y}|\mathbf{x}) \\
&+\mathbb{E}_{q_\phi(\mathbf{y}|\mathbf{x})}[\log p_{\theta}\left(\mathbf{y}|\mathbf{x}\right)].
\end{aligned}
\end{equation}

We summarize the detailed optimization algorithm for \method{} in Algorithm.~\ref{alg::optim}. The model alternates between optimizing the objective function with regard to $\phi$ and $\theta$, which is referred to as E-step and M-step. During each step, both labeled and unlabeled data are leveraged for training. In addition to training two networks with labeled data, we essentially select samples with high confidence from the unlabeled data for regularization to let both networks mutually enhance each other. Specifically, given the unlabeled data $\mathbf{x}$ and their predicted categories $\mathbf{y}$, we rank these data according to the probability of distribution $\mathbf{y}$ (i.e., $q_\phi(\mathbf{y}|\mathbf{x})$ and $p_\theta(\mathbf{y}|\mathbf{x})$) and then select top-$k$ data from the corresponding ranked list. This is similar to the way we do in self-training methods~\cite{lee2013pseudo}. Furthermore, to eliminate the influence of incorrect annotations in selected samples, we collect the intersection of the training samples predicted via both networks and regard them as the additional labeled set to enlarge the labeled data, experimental analysis in the next section shows that consistent prediction data are helpful for improving the final performance. The whole iterative procedure stops until convergence or the unlabeled data are exhausted.

\section{Experiment}
\label{sec::experiment}

To verify the effectiveness of our proposed method, we first introduce the experimental settings and then present the detailed experiment results and performance analysis.

\subsection{Experimental Settings}

\subsubsection{Evaluation Datasets.}
We conduct extensive experiments on seven widely-used semi-supervised graph classification datasets, including two bioinformatics datasets (i.e., PROTEINS~\cite{borgwardt2005protein} and DD~\cite{dobson2003distinguishing}), four social network datasets (i.e., IMDB-B, IMDB-M, REDDIT-B, REDDIT-M-5k~\cite{yanardag2015deep}) and one scientific collaboration dataset (i.e., COLLAB~\cite{yanardag2015deep}). We summarize the statistics of datasets in Table~\ref{tab::datasets_statistics}.
Following InfoGraph~\cite{sun2020infograph}, we use all-ones encoding as input node features when node attributes are not available in the datasets.

\smallskip\noindent\textbf{Data Preparation.}  As these datasets contain labels for all graphs, we need to construct the datasets applicable for semi-supervised learning scenarios. For each dataset, we first split graphs into TRAIN, VAL and TEST sets according to a 7:1:2 ratio. Based on the TRAIN data, we further sample 2/7 of the graphs as labeled data, which is denoted as TRAIN-L and use the rest as unlabeled data TRAIN-U. We use the validation data VAL for hyper-parameter selection and report the results on TEST data set.

\subsubsection{Compared Methods}
We compare our \method{} with the following methods, which can be divided into three categories: traditional graph methods, traditional semi-supervised learning methods and graph-specific semi-supervised learning methods. Traditional graph methods include Graphlet Kernel (GK)~\cite{shervashidze2009efficient}, Shortest Path Kernel (SP)~\cite{borgwardt2005shortest}, Weisfeiler-Lehman Kernel (WL) ~\cite{shervashidze2011weisfeiler}, Deep Graph Kernel (DGK) ~\cite{yanardag2015deep}, Sub2Vec~\cite{adhikari2018sub2vec} and Graph2Vec~\cite{narayanan2017graph2vec}. Traditional semi-supervised learning methods include 
EntMin~\cite{grandvalet2005semi}, $\Pi$-Model~\cite{tarvainen2017mean}, Mean-Teacher~\cite{tarvainen2017mean} and VAT~\cite{miyato2018virtual}. Graph-specific semi-supervised methods include InfoGraph~\cite{sun2020infograph}, ASGN~\cite{hao2020asgn}, GraphCL~\cite{you2020graph} and JOAO~\cite{you2021graph}. Note that we don't compare with SEAL-AI~\cite{li2019semi} because it requires explicit relations among the graph instances, which are not available in our datasets.

\subsubsection{Parameter Settings.}
For the proposed \method{}, we use GIN~\cite{xu2019powerful} to parameterize the GNN-based network $p_\theta$, consisting of three graph convolutional layers and one sum-pooling layer, followed by the softmax function, which is the same as InfoGraph~\cite{sun2020infograph}. The kernel-based network $q_\phi$ is implemented as a memory network with 3 hops, in which the Weisfeiler-Lehman subtree kernel~\cite{shervashidze2011weisfeiler} serves as the base graph kernel. For a fair comparison, we set the batch size to 32 and the number of epochs to 20 for all datasets. The embedding dimensions of hidden layers are set as 32 for bioinformatics datasets while 64 for social network and scientific collaboration datasets. We apply the dropout~\cite{srivastava2014dropout} with a ratio of  $0.5$ in our \method{} model. In each iteration, $p_\theta$ and $q_\phi$ are trained using Adam optimizer \cite{kingma2015adam} where initial learning rate is 0.01 and weight decay is 0.0005.

\begin{table}[t]
\caption{Statistics of the evaluation datasets.}
\centering
\setlength{\tabcolsep}{0.1pt}
\begin{tabular*}{0.45\textwidth}{@{\extracolsep{\fill}}lcccc@{}}
\toprule
\textbf{Datasets}  & Graph Num. & Avg. Nodes  & Avg. Edges & Classes \\ 
\midrule
PROTEINS & 1113  & 39.06  & 72.82 & 2\\
DD & 1178  & 284.32  & 715.66 & 2\\
IMDB-B  & 1000  & 19.77  & 96.53 & 2\\
IMDB-M  & 1500  & 13.00  & 65.94 & 3\\
REDDIT-B & 2000  & 429.63  & 497.75 & 2\\
REDDIT-M-5k & 4999  & 508.52  & 594.87 & 5\\
COLLAB & 5000  & 74.49  & 2457.78 & 3\\
\bottomrule
\end{tabular*}
\label{tab::datasets_statistics}
\end{table}

\begin{table*}[ht]
\caption{Results on seven benchmark datasets. We present the mean together with the standard deviation of prediction accuracy over five runs (in $\%$) using different random seeds. The results in bold indicate the best performance.}
% Our proposed method \method{} outperforms all the baseline methods in most cases.
\label{tab::results}
\centering
\tabcolsep=6.2pt
\begin{tabular}{lccccccc}
\toprule
\multirow{2}*{\bf Methods} &  \multicolumn{6}{c}{\bf Datasets}\\\cmidrule{2-8}%\cline{3-9}
  &  \textbf{PROTEINS} & \textbf{DD}  & \textbf{IMDB-B} & \textbf{IMDB-M}  & \textbf{REDDIT-B} & \textbf{REDDIT-M-5k} & \textbf{COLLAB} \\
\midrule 
GK~\cite{shervashidze2009efficient}  & $ 64.8\pm2.3 $ & $ 53.2\pm1.4 $  & $ 54.5\pm1.7  $ & $  32.3\pm2.4  $  & $  57.8\pm2.7  $  & $ 34.3\pm0.8  $ & $ 55.7\pm1.1  $ \\
SP~\cite{borgwardt2005shortest}  & $ 65.2\pm2.6 $ & $ 55.3\pm2.1 $  & $ 52.0\pm1.6  $ & $  37.7\pm1.9  $  & $  68.3\pm3.7  $  & $ 30.4\pm1.3  $ & $ 64.1\pm1.3  $ \\
WL~\cite{shervashidze2011weisfeiler}  & $ 63.5\pm1.6 $ & $ 57.3\pm1.2 $  & $ 58.1\pm2.3  $ & $  33.3\pm1.4  $  & $  61.8\pm1.3  $  & $ 37.0\pm0.9  $ & $ 62.9\pm0.7  $ \\
DGK~\cite{yanardag2015deep}  & $ 64.4\pm1.7 $ & $ 60.5\pm0.8 $   & $ 55.6\pm2.2  $ & $  34.6\pm1.3  $  & $  66.2\pm2.4  $  & $ 36.5\pm2.4  $ & $ 61.3\pm1.2  $ \\
Sub2Vec~\cite{adhikari2018sub2vec} & $ 52.7\pm4.5 $ & $ 46.4\pm3.2 $   & $ 44.9\pm3.5  $ & $  31.8\pm2.7  $  & $  63.5\pm2.3  $  & $ 35.1\pm1.5  $ & $ 60.8\pm1.4  $ \\
Graph2Vec~\cite{narayanan2017graph2vec}  & $ 63.1\pm1.8 $ & $ 53.7\pm1.6 $  & $ 61.2\pm2.6  $ & $  38.1\pm2.2  $  & $  67.7\pm2.3  $  & $ 38.1\pm1.4  $ & $ 63.6\pm0.9  $ \\
\midrule
EntMin~\cite{grandvalet2005semi} & $ 62.7\pm2.7 $ & $ 59.8\pm1.3 $    & $ 67.1\pm3.7 $ & $ 37.4\pm1.2 $  & $ 66.9\pm3.5 $ & $ 38.7\pm2.8 $ & $ 63.8\pm1.6 $  \\
$\Pi$-Model~\cite{tarvainen2017mean}  & $ 63.2\pm1.2 $ & $ 61.8\pm1.8 $  & $ 67.0\pm3.4 $ & $ 39.0\pm3.5 $  & $ 67.1\pm2.9 $  & $ 39.0\pm1.1 $ & $ 63.7\pm1.0 $ \\
Mean-Teacher~\cite{tarvainen2017mean}  & $ 64.3\pm2.1 $ & $ 60.6\pm1.8 $   & $ 66.4\pm2.7 $ & $ 38.8\pm3.6 $  & $ 68.7\pm1.3 $ & $ 39.2\pm2.1 $  & $ 63.6\pm1.4 $\\
VAT~\cite{miyato2018virtual}  & $ 64.1\pm1.2 $ & $ 59.9\pm2.6 $   & $ 67.2\pm2.9 $ & $ 39.6\pm1.4 $  & $ 70.8\pm4.1 $  & $ 38.9\pm3.2 $ & $ 64.1\pm1.1 $ \\
\midrule
InfoGraph~\cite{sun2020infograph}  & $ 68.2\pm0.7 $& $ 67.5\pm1.4 $  & $ 71.8\pm2.3 $ &  $ 42.3\pm1.8 $  & $ 75.2\pm2.4 $ & $ 41.5\pm1.7 $ & $ 65.7\pm0.4 $ \\
ASGN~\cite{hao2020asgn}  & $ 67.7\pm1.2 $ & $ 68.5\pm0.6 $  & $ 70.6\pm1.4 $ & $ 41.2\pm1.4 $  & $ 73.1\pm2.3 $ & $ 42.2\pm0.8 $ & $ 65.3\pm0.8 $\\
GraphCL~\cite{you2020graph} & $ 69.4\pm0.8 $ & $ 68.7\pm1.2 $ & $ 71.2\pm2.5 $ & \textbf{43.7 $\pm$ 1.3} & $ 75.2\pm1.7 $ & $ 42.3\pm0.9 $ & $ 66.4\pm0.6 $\\
JOAO~\cite{you2021graph} & $ 68.7\pm0.9 $ & $ 67.9\pm1.3 $ & $ 71.0\pm1.9 $ & $ 42.6\pm1.5 $ & $ 74.8\pm1.6 $ & $ 42.1\pm1.2 $ & $ 65.8\pm0.4 $\\
\midrule 
\method{} (Ours)  &  \textbf{70.9 $\pm$ 0.5}  &  \textbf{70.5 $\pm$ 0.6}   &  \textbf{72.5 $\pm$ 1.6}  &  43.3 $\pm$ 0.7     &  \textbf{76.0 $\pm$ 0.9}  &  \textbf{44.8 $\pm$ 0.6}  &  \textbf{67.4 $\pm$ 0.5} \\
\bottomrule
\end{tabular}
\end{table*}

\subsection{Results}
We report the quantitative results of semi-supervised graph classification with half of the labeled data TRAIN-L in Table~\ref{tab::results}. From this table, we make the following observations.
(1) We can observe that most of the traditional graph methods perform worse than other methods, which shows that graph neural networks possess the strong representation-learning ability to extract more useful information from graph-structured data. 
(2) The methods with traditional semi-supervised learning techniques (EntMin, $\Pi$-Model, Mean-Teacher and VAT) show worse performance compared with the graph-specific semi-supervised learning approaches, indicating that recent graph semi-supervised learning techniques are more suitable for the challenging graph classification task. 
(3) Among the previous state-of-the-art approaches, GraphCL achieves almost the best performance on most datasets. Specifically, other graph-specific semi-supervised learning approaches are not as effective as GraphCL, maybe the reason is that GraphCL takes advantage of the idea of instance discrimination for contrastive learning and specific graph augmentation strategies, and thus further improve the performance.
(4) Our framework \method{} achieves the best performance on six of seven datasets, which demonstrates the effectiveness of our framework. From the results, we attribute the performance gain to two aspects: (i) Introduction of graph kernels. Graph kernels acting as structured constraints are able to guide the training process of graph neural networks. (ii) Introduction of posterior regularization framework. Instead of simply integrating the graph neural networks and graph kernels, we use a principled way to combine both worlds, which can mutually enhance each other. 

\begin{figure}[t]\small
\centering
\subfigure[\textbf{PROTEINS}]{
\centering
\includegraphics[width=0.23\textwidth]{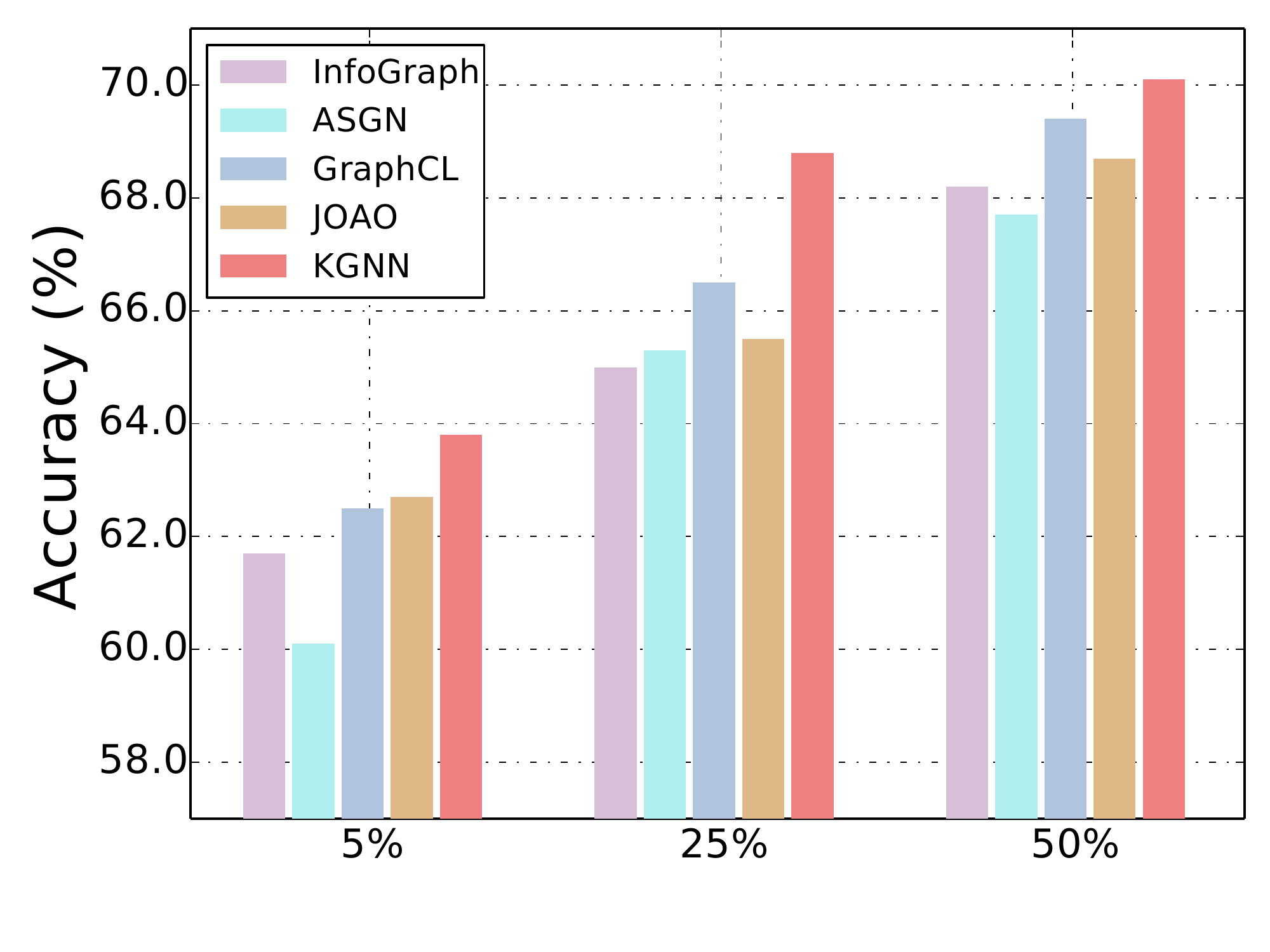}
}
\hspace{-2mm}
\subfigure[\textbf{IMDB-B}]{
\centering
\includegraphics[width=0.23\textwidth]{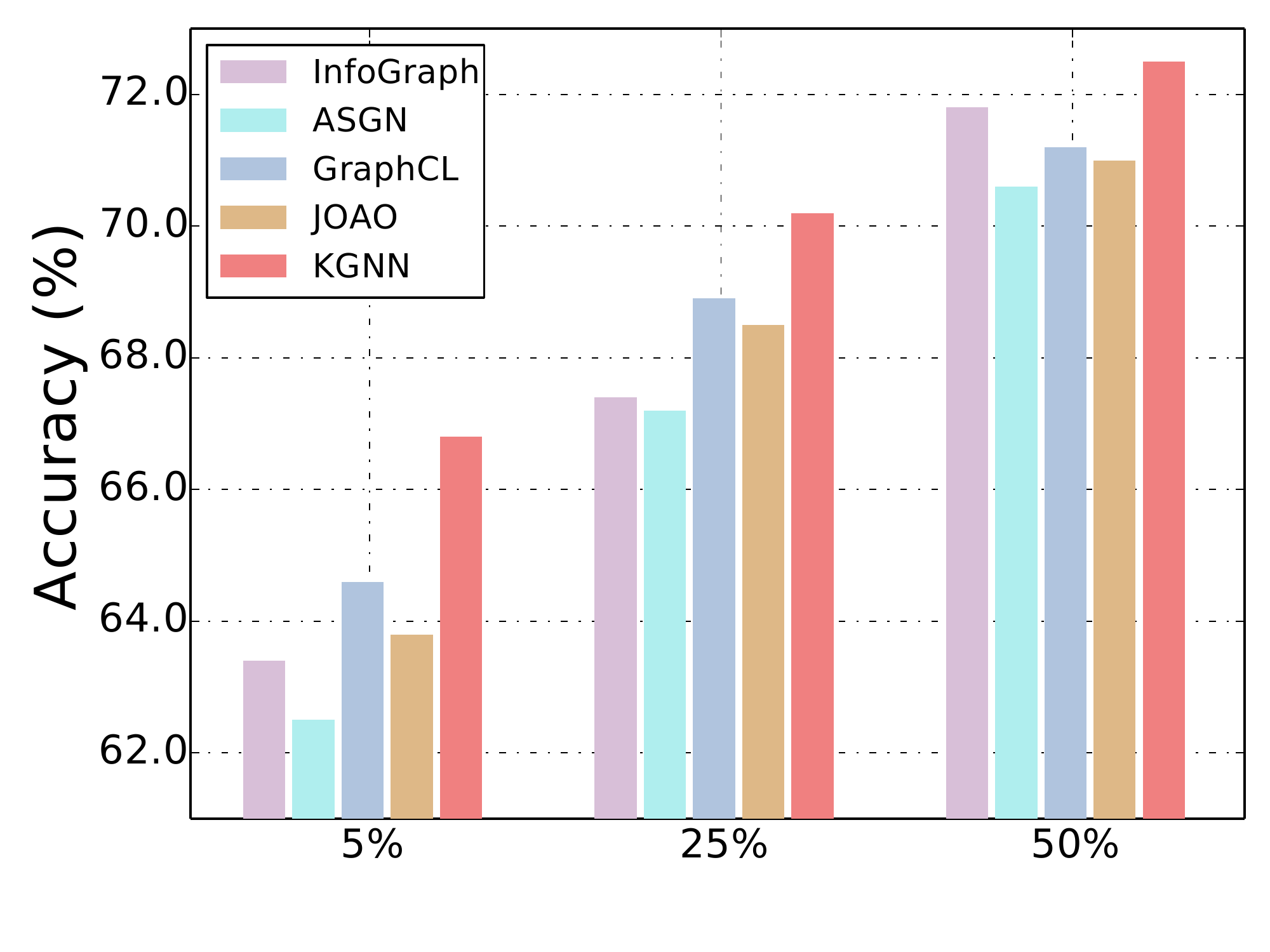}
}
\vspace{-2mm}
\caption{Results on two datasets \textit{w.r.t.} the amounts of the labeled data (i.e., $5\%$, $25\%$ and $50\%$) and all the unlabeled data.}
\label{fig::label_ratio}
\end{figure}

\noindent\textbf{Performance on Different Amounts of Labeled Data.} We vary rates of the labeled data for training to show the performance of different models in Figure \ref{fig::label_ratio}. We take the PROTEINS and IMDB-B as examples. Here we omit the traditional methods since they do not have competitive performance. From the results, we find that the performance of all the methods further improves as the number of available labeled data increases, demonstrating that adding more labeled data is an efficient approach to boost the performance. Among all the methods, the proposed \method{} achieves the best performance, which indicates that explicitly incorporating graph kernels as structured constraints into graph neural networks can further facilitate the performance for graph classification.

\subsection{Ablation Study}
In the \method{} model, either the GNN-based network or the kernel-based network is enhanced by incorporating extra ``labels'' selected from both networks. A more straightforward way could be empowering every single network with ``labels'' generated by itself, which is known as the self-training method. Next, we investigate a few variants that are trained under the self-training framework:

\begin{itemize}[leftmargin=*]
    \item \textbf{GNN-Sup.} Our base model, which trains a GNN (i.e., $p_\theta$ network) solely on labeled data TRAIN-L in a fully supervised manner. 
    \item \textbf{MemNN-Sup.} It is another model, which optimizes a MemNN (i.e., $q_\phi$ network) solely on labeled data in a supervised way. 
    \item \textbf{GNN-Self.} It is a model variant where we optimize a GNN (i.e., $p_\theta$) using a self-training strategy.
    \item \textbf{Ensemble-Self.} Besides using a single network, we first ensemble the GNN-based network $p_\theta$ and the kernel-based network $q_\phi$ by concatenating their graph representations before the final prediction layer, and train the ensembled model with self-training.  
    \item \textbf{KGNN-Sep.} KGNN-Sep differs from the original KGNN in that we directly feed the label annotations by one network into the other network without checking the prediction consistency.
\end{itemize}

\begin{table*}[ht]
\renewcommand\arraystretch{0.9}
\caption{Comparison with several variants for ablation study (in $\%$). We present the mean together with the standard deviation of prediction accuracy over five runs (in $\%$) using different random seeds. The results in bold indicate the best performance.}
\label{tab::ablation}
\centering
\tabcolsep=7.3pt
\begin{tabular}{lccccccc}
\toprule 
\multirow{2}*{\bf Methods}   &  \multicolumn{7}{c}{\bf Datasets}  \\\cmidrule{2-8}%\cline{2-8}
& \textbf{PROTEINS} & \textbf{DD}  & \textbf{IMDB-B} & \textbf{IMDB-M}  & \textbf{REDDIT-B} & \textbf{REDDIT-M-5k} & \textbf{COLLAB} \\
\midrule 
GNN-Sup  & $ 63.3\pm1.4 $ & $ 62.5\pm1.5 $  & $ 63.4\pm2.1 $ & $ 39.2\pm1.6 $ & $ 69.8\pm1.1 $ & $ 38.6\pm2.5 $ & $ 61.7\pm1.5 $\\
MemNN-Sup  & $ 59.8\pm2.4 $ & $ 60.4\pm1.0 $  & $ 59.3\pm2.7 $ & $ 38.3\pm2.8 $  & $ 65.2\pm1.2 $  & $ 39.0\pm2.6 $ & $ 60.1\pm2.2 $ \\
GNN-Self   & $ 65.2\pm1.6 $ & $ 64.0\pm1.4 $  & $ 67.3\pm1.8 $ & $ 41.1\pm2.2 $ & $ 68.3\pm1.8 $ & $ 42.1\pm2.1 $ & $ 63.5\pm1.6 $  \\
Ensemble-Self  & $ 65.5\pm2.4 $ & $ 66.1\pm3.4 $  & $ 68.0\pm2.8 $ & $ 41.0\pm3.1 $  & $ 72.4\pm0.3 $  & $ 41.5\pm2.8 $ & $ 64.3\pm1.0 $ \\
\method{}-Sep  & $ 68.8\pm2.9 $ & $ 67.4\pm1.5 $  & $ 68.8\pm2.2 $ & $ 42.4\pm3.2 $ & $ 73.6\pm2.7 $ & $ 42.8\pm1.3 $ & $ 65.1\pm1.2 $ \\
\midrule 
\specialrule{0em}{1pt}{1pt}
\method{} (Ours)  &  \textbf{70.9 $\pm$ 0.5}  &  \textbf{70.5 $\pm$ 0.6}   &  \textbf{72.5 $\pm$ 1.6}  &  \textbf{43.3 $\pm$ 0.7}     &  \textbf{76.0 $\pm$ 0.9}  &  \textbf{44.8 $\pm$ 0.6}  &  \textbf{67.4 $\pm$ 0.5} \\

\bottomrule
\end{tabular}
\end{table*}

\noindent\textbf{Results and Analysis.} The results of the above model variants are summarized in Table~\ref{tab::ablation}. We can see that GNN-Sup outperforms MemNN-Sup on most datasets, maybe the reason is that GNN-based methods can take advantage of node attributes while graph kernel-based methods fail. Also, GNN-Self outperforms GNN-Sup on 5 out of 7 datasets. On DD and REDDIT-B datasets, GNN-self performs worse than GNN-Sup, which may attribute to the noisy ``labels'' generated by the model itself. Importantly, we find that Ensemble-Self performs well compared with the above three model variants in most cases, which implies that incorporating graph similarity explicitly indeed benefits the performance. Besides, by using a more principled way to combine two networks, KGNN-Sep outperforms the above four models on all datasets, showing the effectiveness of our proposed posterior regularization. Finally, with the intersection strategy to eliminate the influence of incorrect annotations, our proposed \method{} achieves the best performance on all datasets, which is in accordance with our expectations.

\subsection{Parameter Analysis}
Finally, we examine the \method{}'s performance varies w.r.t. different parameter settings. Specifically, we investigate the effect of embedding dimensions of hidden layers and the number of hops in the memory network on two datasets PROTEINS and COLLAB.

\begin{figure} 
% 	\hspace{-0.03\linewidth}
	\begin{minipage}[t]{0.48\linewidth} 
	\centering 
	\includegraphics[width=\textwidth]{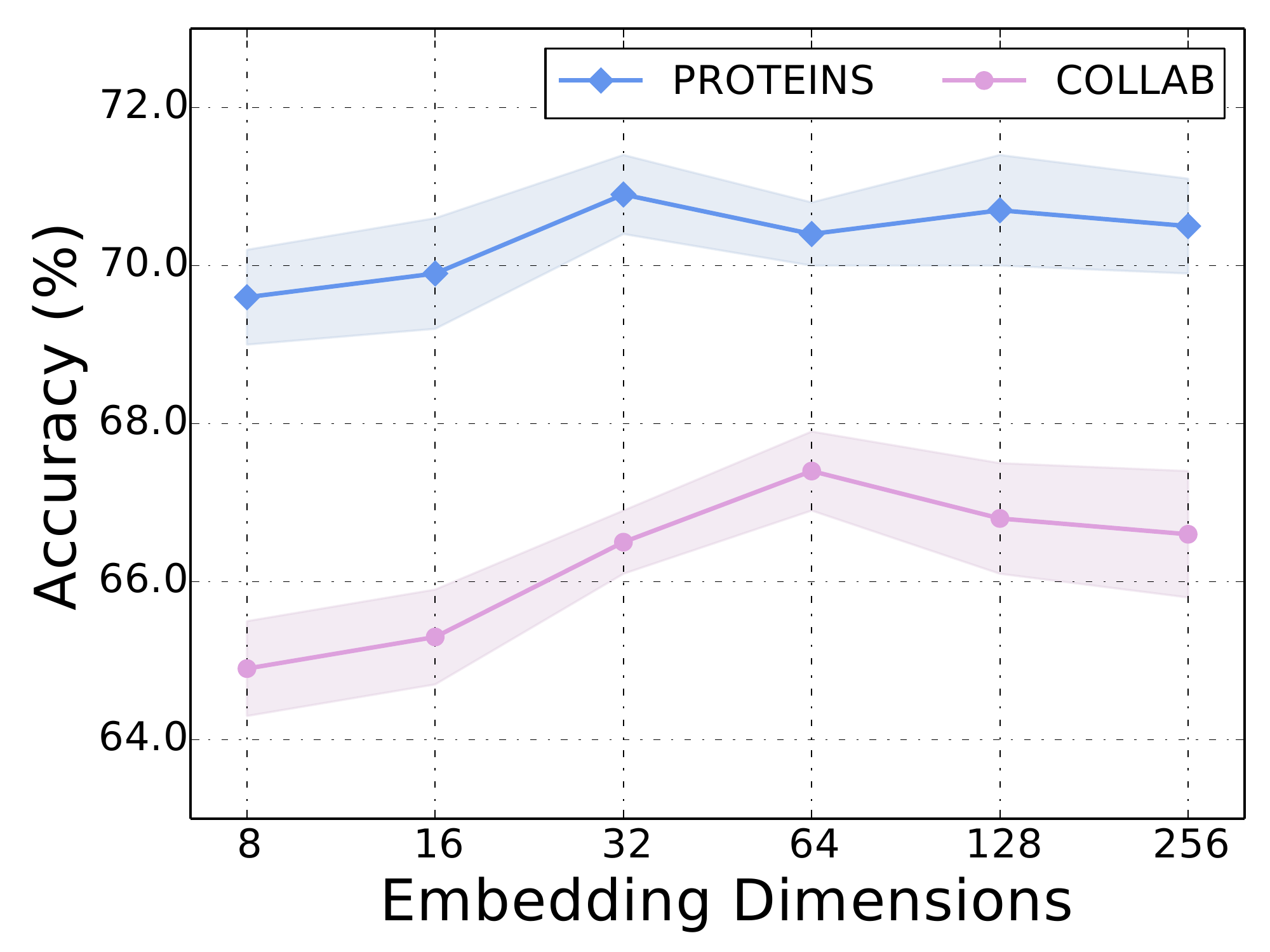} 
	\caption{Performance \textit{w.r.t.} the embedding dimensions.} 	
	\label{fig::parameter_dim} 
	\end{minipage}% 
	\hspace{0.01\linewidth}
	\begin{minipage}[t]{0.48\linewidth} 
	\centering 
	\includegraphics[width=\textwidth]{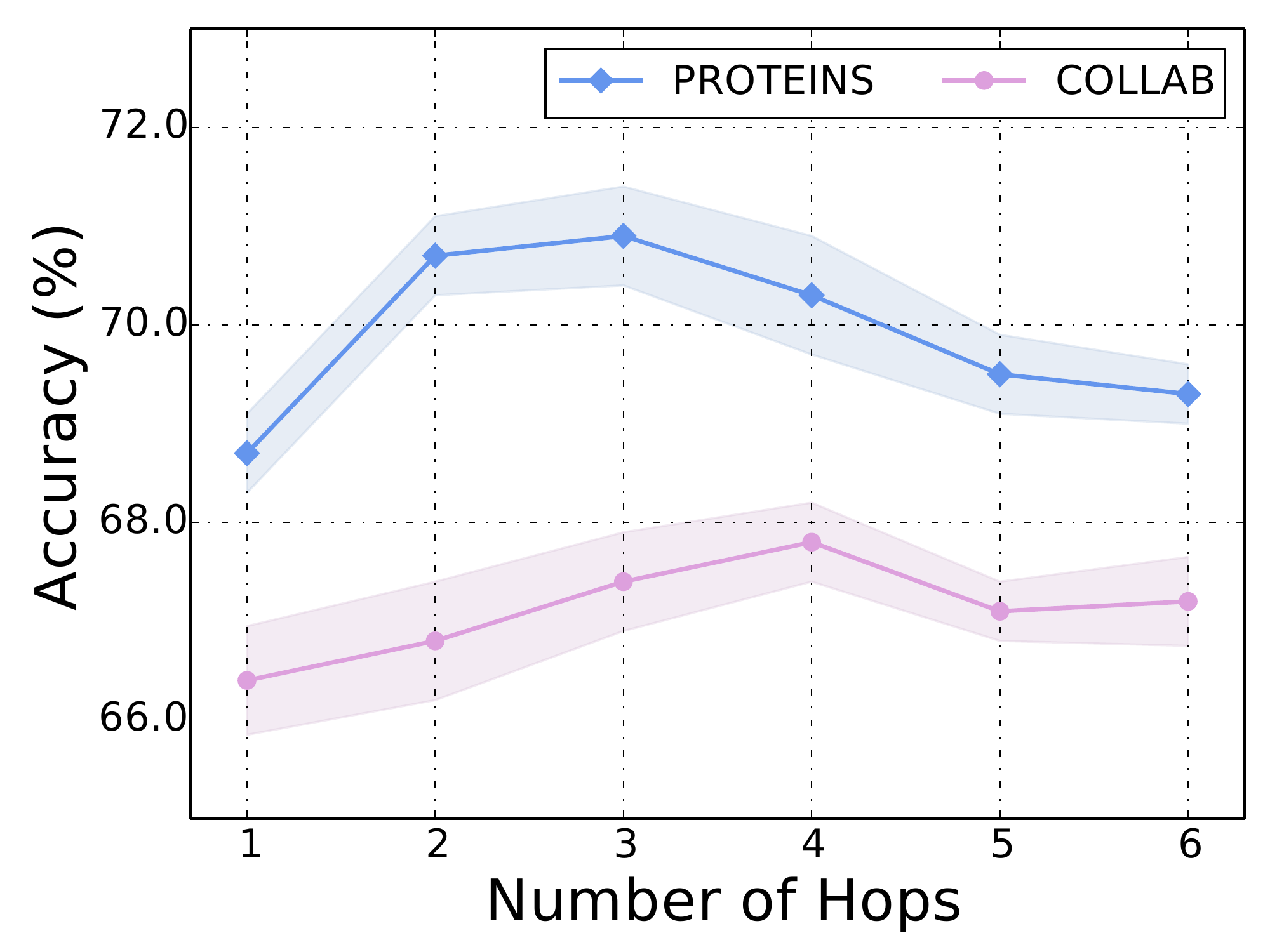} 
	\caption{Performance \textit{w.r.t.} the number of hops.} 
	\label{fig::parameter_hops} 
	\end{minipage}% 
\end{figure}

\smallskip
\noindent\textbf{Effect of the Embedding Dimensions.}
We first analyze the effect of the embedding dimensions of hidden layers $h$. We expect the model to perform well as dimensions increase due to the benefit of augmented model capacity. We fix all other parameters to the ones that yield the best results and varied $h = \{8, 16, 32, 64, 128\}$, the obtained results are shown in Figure~\ref{fig::parameter_dim}. We observe that larger embedding dimensions generally lead to better performance before saturation. The model needs larger embedding dimensions on COLLAB than PROTEINS to achieve the best performance as its dataset is large-scale and requires a larger capacity model to fit.

\smallskip
\noindent\textbf{Effect of the Number of Hops.}
We further assess the effect of the number of hops in the memory network. We vary the number of hops from 1 to 6 while fixing all other parameters. As shown in Figure~\ref{fig::parameter_hops}, we observe that increasing the number of hops results in better performance when the number is small, showing that the multi-hop mechanism has more broad attention and helps in learning correlations between memories. However, too many hops may hurt the performance, maybe the reason is that the query graph pays much attention to how similar it is with the in-memory graphs, which leads to overfitting and loses the generalization of the comparison similarity with the unknown graphs.

\subsection{Case Study}

\begin{figure}[t]
	\centering
	\includegraphics[width=1.0\linewidth]{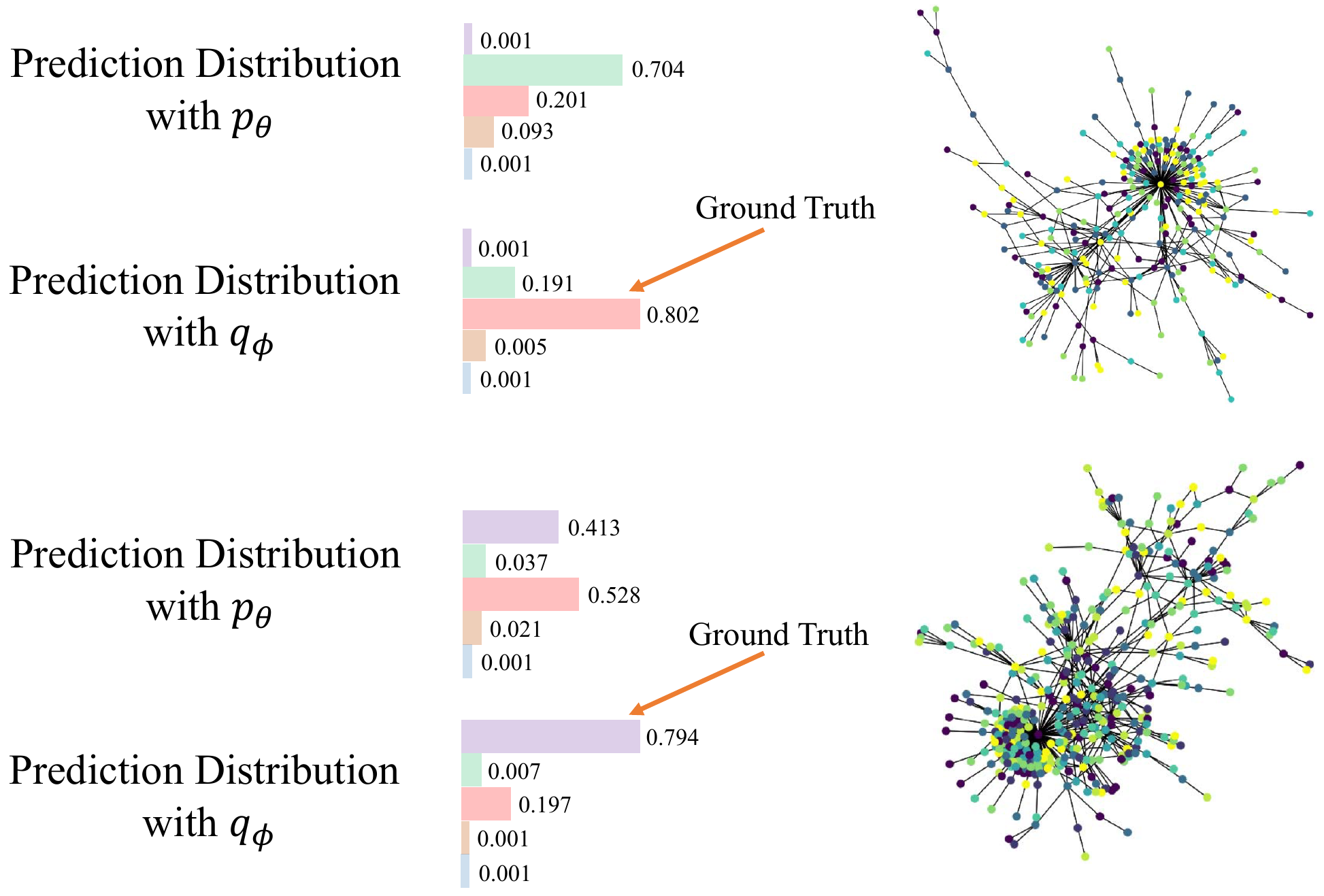}
	\caption{Illustration of two samples on REDDIT-M-5k.}
	\label{fig::case_study}
\end{figure}

In this section, we study the power of the kernel-based network to validate the strength of our framework, showing the superiority from the view of topological similarity. Figure \ref{fig::case_study} illustrates some of the results we obtained on REDDIT-M-5k, where two unlabeled samples are annotated with both GNN-based network $p_\theta$ and kernel-based network $q_\phi$. We observe that both samples are classified into wrong categories by $p_\theta$ while $q_\phi$ corrects the wrong prediction successfully, which shows that our novel kernel-based network is a necessary supplement for the effective graph classification task. The reason may be that the kernel-based network is able to capture more relationships between different substructures and thus acts as structured constraints to guide the GNN-based network. 

\section{Conclusion}
\label{sec::conclusion}
This paper studies semi-supervised graph classification, which is a fundamental problem in graph-structured data modeling, and a novel approach called the \method{} is proposed. \method{} adopts the posterior regularization to incorporate graph kernels as structured constraints to guide the training process of graph neural networks under the EM-style framework. Specifically, in the M-step, we employ a GNN-based network, which can learn effective graph representations for prediction. In the E-step, we develop a kernel-based network using a memory network to capture the graph similarity efficiently.
Experiments on a range of well-known benchmark datasets prove the effectiveness of the \method{} for graph classification. In the future, we plan to further extend our \method{} to a broader range of applications, such as drug discovery and material science.

\begin{acks}
This paper is partially supported by National Key Research and Development Program of China with Grant No. 2018AAA0101902 as well as the National Natural Science Foundation of China (NSFC Grant No. 62106008 and No. 62006004). 
\end{acks}


\balance
\bibliographystyle{ACM-Reference-Format}
\bibliography{7-ref}

\end{document}